\pdfoutput=1

\documentclass[11pt]{article}

\usepackage[preprint]{acl}

\usepackage{times}
\usepackage{latexsym}

\usepackage[T1]{fontenc}

\usepackage[utf8]{inputenc}

\usepackage{microtype}

\usepackage{inconsolata}

\usepackage{hyperref}       
\usepackage{url}            
\usepackage{booktabs}       
\usepackage{amsfonts}       
\usepackage{amsmath}
\usepackage{nicefrac}       
\usepackage{microtype}      
\usepackage{pifont}
\usepackage[table,xcdraw,dvipsnames]{xcolor}

\usepackage{lipsum}
\usepackage{footnote}
\usepackage{todonotes}
\usepackage{colortbl}

\usepackage{xspace}
\usepackage{xparse}
\usepackage{graphicx}
\usepackage{ragged2e}

\usepackage{enumitem}

\usepackage{array}
\newcolumntype{C}[1]{>{\centering\arraybackslash}p{#1}}

\NewDocumentCommand{\rot}{O{45} O{1em} m}{\makebox[#2][l]{\rotatebox{#1}{#3}}}%
\NewDocumentCommand{\rotb}{O{-45} O{1em} m}{\makebox[#2][l]{\rotatebox{#1}{#3}}}%

\newcommand{\matteroffact}{\textsc{Matter-of-Fact}\xspace}

\title{\textsc{Matter-of-Fact:} A Benchmark for Verifying the Feasibility of \\Literature-Supported Claims in Materials Science} 

\author{Peter Jansen \\\And
  Samiah Hassan \\\And
  Ruoyao Wang \\
  \texttt{pajansen@arizona.edu}
  }

\author{
  \textbf{Peter Jansen\textsuperscript{1,2}},
  \textbf{Samiah Hassan\textsuperscript{1}},
  \textbf{Ruoyao Wang\textsuperscript{1}}
\\
  \textsuperscript{1}University of Arizona,
  \textsuperscript{2}Allen Institute for Artificial Intelligence
\\
\texttt{pajansen@arizona.edu}
}

\begin{document}
\maketitle
\begin{abstract}
Contemporary approaches to assisted scientific discovery use language models to automatically generate large numbers of potential hypothesis to test, while also automatically generating code-based experiments to test those hypotheses.  While hypotheses can be comparatively inexpensive to generate, automated experiments can be costly, particularly when run at scale (i.e. thousands of experiments).  Developing the capacity to filter hypotheses based on their feasibility would allow discovery systems to run at scale, while increasing their likelihood of making significant discoveries. In this work we introduce \matteroffact, a challenge dataset for determining the feasibility of hypotheses framed as claims, while operationalizing feasibility assessment as a temporally-filtered claim verification task using backtesting. \matteroffact includes \textsc{8.4k} claims extracted from scientific articles spanning four high-impact contemporary materials science topics, including superconductors, semiconductors, batteries, and aerospace materials, while including qualitative and quantitative claims from theoretical, experimental, and code/simulation results. We show that strong baselines that include retrieval augmented generation over scientific literature and code generation fail to exceed 72\% performance on this task (chance performance is 50\%), while domain-expert verification suggests nearly all are solvable -- highlighting both the difficulty of this task for current models, and the potential to accelerate scientific discovery by making near-term progress.\footnote{Benchmark, models, and claim extraction system: ~~~\url{https://github.com/cognitiveailab/matter-of-fact}}
\end{abstract}

\section{Introduction}

Contemporary language models are being broadly integrated into the scientific discovery pipeline.  Existing systems can generate hypothesis \cite{si2024can,radensky2024scideator}, run experiments \cite{lu2024ai,li2024mlr,jansen2025codescientist}, analyze data \cite{majumder2025discoverybench}, and write or review papers \cite{liu2023reviewergpt,zhou2024llm}.  A central benefit -- and challenge -- of these systems is that they can function at scales greater than any human scientist.  For example, hypothesis generation systems might easily produce thousands of potential hypotheses \cite{lu2024ai,jansen2025codescientist}, and running experiments to test each of these would be costly and impractical -- particularly in that few experiments are likely to yield positive results.  In this work we investigate the task of \textit{feasibility assessment} \citep[e.g.][]{o2025sparks}, or assessing whether we can filter hypothesis (expressed as claims) to those that are most likely to be feasible, and have their hypotheses confirmed.  Performing well at this task would allow us to incorporate feasibility filtering in hypothesis generation systems, and potentially make more discoveries with a given (fixed) experimental budget. 

Feasibility assessment is in principle quite challenging as it involves (at times) a high degree of uncertainty in predicting future results, and yet it is a task that scientists perform frequently during experiment planning stages -- selecting the hypotheses that we believe are most likely to return positive results based on a combination of literature, pilot experiments or analyses (which may include empirical work, or code/simulations), and past experience.  In this work we aim to investigate how well current models can perform this feasibility assessment task, and provide a benchmark to assist in improving model performance over time. 

%
%
\begin{figure*}[t]
  \centering
  \includegraphics[scale=1.07]{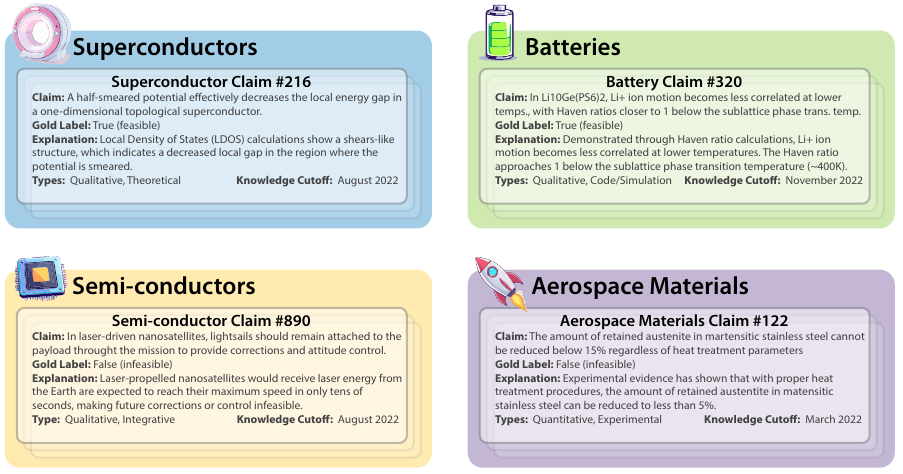}
  \caption{\footnotesize Examples of the four main materials science topics included in \matteroffact, including \textit{superconductors}, \textit{semi-conductors}, \textit{batteries}, and \textit{aerospace materials}.  Each claim includes the claim text, gold label \textit{(true/feasible} or \textit{false/infeasible}), a brief explanation and supporting facts based from the original paper the claim was sourced from, additional meta-data (such as whether the claim is \textit{quantitative} or \textit{qualitative}), and the knowledge cutoff date for the feasibility assessment task.}
  \label{fig:example_data}
\end{figure*}

Generating data to test feasibility assessment is challenging, as (by nature) the experimental results of proposed hypotheses are as yet unknown, which makes gold labels for determining whether a hypothesis is feasible or infeasible effectively unavailable.  To address this challenge, \textit{we operationalize feasibility assessment as a temporally-filtered claim verification task} using the concept of backtesting.  As with conventional claim verification tasks \citep[e.g.][]{thorne-etal-2018-fever}, we generate a corpus of claims extracted from recent scientific literature -- however, in addition, each claim is paired with a ``knowledge cut-off date'', which is the date that the paper the claim was generated from was first authored.  When performing the feasibility assessment task, models are allowed to use any information available before the source paper was authored to assess feasibility, essentially rewinding into the past to predict future results.  In this way, models are provided with knowledge up to (for example) 2023, and must use that knowledge (through a combination of literature search, small-scale code-based experimentation, world modeling, or other methods) to predict whether genuine results (and artificially-generated infeasible results) from 2024 onward are feasible or infeasible. 

The contributions of this work are: 
\begin{enumerate}
    \item We introduce \matteroffact, a benchmark of \textsc{8.4k} claims extracted from recent materials science articles in four high-impact subdomains.  Each claim includes categorical information (qualitative vs quantitative, and experiment, code, or theory focused), and is paired with a knowledge cut-off date to use for the feasibility assessment task. 
    \item We empirically demonstrate that strong baseline models using a variety of solution methods (including retrieval-augmented generation with \textsc{SemanticScholar}, as well as evidence gathered from code-generation) across base models (\textsc{gpt-4o-mini}, \textsc{o4-mini}, and \textsc{Claude Sonnet 3.7}) achieve a maximum of 72\% accuracy, highlighting the challenging nature of this feasibility assessment task.
    \item We assess the quality of the claims both by domain expert evaluation, and by evaluating base models in a conventional claim verification task. Humans and models reach 93\%+, suggesting the benchmark is of high quality. 
\end{enumerate}

%
%
\begin{table*}[t!]
\centering
\footnotesize
{\setlength{\tabcolsep}{5pt}
\begin{tabular}{lccccccc}
\toprule
\textbf{Benchmark}                              &   \textbf{Domain}           & \textbf{Claims}  &   \textbf{Source}   &   \textbf{Generation Method}      \\
\midrule
\textsc{SciFact} \cite{wadden-etal-2020-fact}            &  Biomed          &   \textsc{1.4k}    &   Paper Abstracts   & Citances provided to annotators  \\
\textsc{COVID-Fact} \cite{saakyan-etal-2021-covid}         &  Biomed        &   \textsc{4.0k}    &   Reddit            & Extract positive, generate counterclaim  \\
\textsc{SciFact-Open} \cite{wadden-etal-2022-scifact}       &  Biomed       &   \textsc{1.4k}    &   Paper Abstracts   & See \textsc{SciFact}  \\
\textsc{ClaimCheck} \cite{Ou2025CLAIMCHECKHG}         &  ML                 &   \textsc{154}     &   Paper Reviews     & Emphasizes claim weaknesses  \\
\textsc{SciTab} \cite{lu-etal-2023-scitab}             &  Comp. Sci         &   \textsc{1.2k}    &   Paper Tables      & Compositional reasoning on tables  \\
\midrule
\textbf{\textsc{Matter-Of-Fact} (This work)}             &  \textbf{Mat. Sci} &   \textbf{\textsc{8.4k}}    &   \textbf{Paper full-text}   & \textbf{Extract positive, generate infeasible}  \\
\bottomrule        
\end{tabular}
}
\caption{\footnotesize A comparison of claim verification datasets with \matteroffact, including their domain, size, source of the information used to generate or extract claims from, and the claim generation method.\label{tab:env_comparison}}
\end{table*}

\section{Related Work}
{\flushleft{\textbf{Scientific Claim Verification Datasets:}}} The scientific claim verification task requires a model to determine whether a claim (typically extracted from a scientific paper) is true or false, either by leveraging its pretrained scientific knowledge or retrieving evidence from a corpus, with a selection of scientific claim verification benchmarks shown in Table~\ref{tab:env_comparison}. \textsc{SciFact} \cite{wadden-etal-2020-fact} contains \textsc{1.4K} biomedical-domain claims generated by showing citances (sentences that cite a paper and describe its contribution) to human annotators, who were then asked to generate associated claims.
Where \textsc{SciFact} pairs claims with a set of \textsc{5k} abstracts that can be used for gathering evidence, \textsc{SciFact-Open} \cite{wadden-etal-2022-scifact} expands this evidence retrieval corpus to \textsc{500K} abstracts, presenting a more challenging retrieval problem.
Also in the biomedical domain, \textsc{COVID-Fact}~\cite{saakyan-etal-2021-covid} consists of over \textsc{4K} claims extracted from Reddit. Lu et al.~\shortcite{lu-etal-2023-scitab} introduce \textsc{SciTab}, which requires verifying computer science claims centrally using tables extracted from papers.  \textsc{ClaimCheck} \cite{Ou2025CLAIMCHECKHG} uses reviews of rejected NeurIPS submissions from OpenReview to build a corpus of 154 claims that emphasize identifying the weaknesses in scholarly claims.  In contrast, \matteroffact builds a corpus of \textsc{8.4k} materials science claims for feasibility assessment that are generated from the nuanced results found in the full text of source articles (rather than abstracts), and where negative claims focus on being scientifically infeasible rather than factually incorrect.

{\flushleft{\textbf{Claim Verification Models:}}} Our framing of feasibility detection is as temporally-filtered claim verification with a knowledge cutoff.  More broadly, recent approaches to claim verification typically involve two key steps: evidence retrieval and fact checking. For the retrieval step, augmenting LLMs with retrieved documents \cite{Izacard2022FewshotLW} or knowledge bases~\cite{baek-etal-2023-knowledge-augmented-language, 10480162} can be effective for improving fact verification performance of models. 
$\text{Re}^2\text{G}$~\cite{glass-etal-2022-re2g} extends the retrieval step with a trained reranker to achieve better retrieval performance for fact checking. 
Rani et al.~\shortcite{rani-etal-2023-factify} propose a form of query expansion that generates claim-related questions as queries to retrieve supporting documents. For fact checking, some methods make use of structured knowledge representations such as knowledge graphs \cite{dammu-etal-2024-claimver} and first-order-logic~\cite{wang-shu-2023-explainable} to organize evidence and verify facts. End-to-end systems combine the entire retrieval and verification pipeline, such as \textsc{ARSJoint} \cite{zhang-etal-2021-abstract} and \textsc{SciClaims }\cite{ortega2025sciclaimsendtoendgenerativebiomedical}. In this work we demonstrate similar retrieval-backed systems (with temporal filtering) for feasibility assessment, while also providing formal approaches based on code generation.

{\flushleft{\textbf{Scientific Discovery and Feasibility Assessment:}}} Automated scientific discovery is frequently divided into two subfields: problem-specific methods (like AlphaFold~\cite{Jumper2021HighlyAP} for protein structure prediction), and problem-general methods that work across a variety of problem types.  Examples of problem-specific systems in the materials science domain include GNoME~\cite{Merchant2023}, a graph neural network (GNN) based method that discovered over 2.2 million new stable crystal structures, and Schmidt et al.~\shortcite{Schmidt2023MachineLearningAssistedDO}'s method for using crystal-graph neural networks together with high-quality data for accurate stability prediction. The latter work screened 1 billion materials, discovering 150k+ stable compounds, and identified extreme-property materials like superconductors.  Similarly, Chen et al.~\shortcite{Chen2024AcceleratingCM} combine machine learning models with traditional physics-based models to discover compounds to which can potentially serve as solid electrolytes. These problem-specific methods can be applied to feasibility assessment by predicting highly specific properties of unknown materials. 
\matteroffact works to bridge the gap between problem-specific methods and problem-general methods by providing a large set of claims across 4 broad and high-impact areas of materials science, each of which is likely to benefit from a variety of problem-specific methods to arrive at accurate feasibility assessments. As we empirically demonstrate in our modeling results, because \matteroffact nominally requires a large set of capacities to solve, it is challenging benchmark for measuring a general capacity to assess feasibility over broad subdomains.

\section{Dataset}
\label{sec:dataset}

The \textsc{Matter-of-Fact} benchmark consists of \textsc{8.4k} claims extracted from the full-text of materials science articles.   The extraction and validation process is described below, with example claims shown in Figure~\ref{fig:example_data}.

{\flushleft\textbf{Inclusion Criteria:}} We assembled a corpus of recent publicly-available materials science domain articles by crawling Arxiv for all papers within the \textsc{materials science} and \textsc{superconductivity} topics submitted on or after January 2022, resulting in a total of \textsc{24k} articles.  Articles were then filtered based on specific inclusion criteria.  First, articles that were not licensed using a specific permissive license (\textsc{Creative Commons-By Attribution-4.0}) were removed. Second, to prevent having to use a \textsc{PDF-to-text} conversion pipeline (which can have limited quality on complex tables, chemical formulas, mathematics, and other artifacts found within materials science articles), we further filtered to include only articles with \LaTeX ~source available.  Papers with long source ($>$30k tokens) were also removed (approximately 16\% of articles). After initial filtering, \textsc{4.2k} articles remained.  Our focus in this work is specifically in four high-impact subdomains: superconductors, semi-conductors, batteries, and aerospace materials.  To identify articles within these topics, we performed topic labeling of each abstract using \textsc{gpt-4o-mini} with a prompt that emphasized identifying articles within these 4 focus areas. We then sampled 500 total articles (125 from each topic) to use for claim generation.  

{\flushleft\textbf{Initial Claim Generation:}} Claims were generated by providing the full-text (\LaTeX ~source) of each paper in a prompt, together with task instructions and \textsc{JSON} output format requirements.  The model was instructed to generate matched pairs of claims -- one true, and one that was clearly false or infeasible -- and for each claim, to provide a list of supporting evidence, followed by an overall explanation as to why the evidence supports or refutes the claim.\footnote{Scientific articles tend to express positive claims rather than negative claims. We follow the approach of Saakyan et al.~\shortcite{saakyan-etal-2021-covid} to first extract positive claims, then automatically generate negative claims from these positive references.} 

Claims were instructed to be stand-alone, and not make reference to the paper in the claim text (i.e. ``\textit{Table 4} claims the boiling point of Material X is...''), so that they could be (in principle) solved without reference to the original source paper.  Negative claims were instructed to be false or clearly infeasible (but not overly so), and not simply claims for which no evidence was available.  Similarly, negative claims were instructed to use balanced language so as not to give away their true or false nature by particulars of wording, such as through use of negation markers (i.e. ``Material X does \textit{not} have...''), and to instead use neutral framings. 
In addition to the above constraints, claims were explicitly asked to be authored on two dimensions.  The first asks for claims to specifically test either \textit{qualitative} knowledge (e.g. ``\textit{In Situation X, Phenonemon Y helps Material Z maintain its superconductivity}''), or \textit{quantitative} knowledge (e.g. ``\textit{Material X superconducts at 77 Kelvin}'').  Second, claims were asked to be authored cross 4 main types: those that focus on \textit{experimental results}, \textit{code/simulation results}, \textit{theoretical results}, or \textit{integrative methods} across types.

{\flushleft\textbf{Balanced Temporal Sets:}} Claims were temporally sorted into those from papers first appearing on Arxiv in 2022 (for training), those in 2023 (for validation), and the most recent claims from papers submitted between 2024 and April 2025 (for testing).  For each set, we filtered claims such that equal numbers of true and false claims were present, to achieve a baseline (random chance) performance of 50\%.  Claims were also balanced such that equal numbers within the \textit{experimental, code, theory,} and \textit{integrative} categories appear within a given set. The final dataset includes a total of \textsc{8.4k} claims, distributed as \textsc{1.4k} claims for training, \textsc{2.5k} for validation, and \textsc{4.4k} for testing. 

%
%
\begin{figure*}[t]
  \centering
  \includegraphics[scale=1.46]{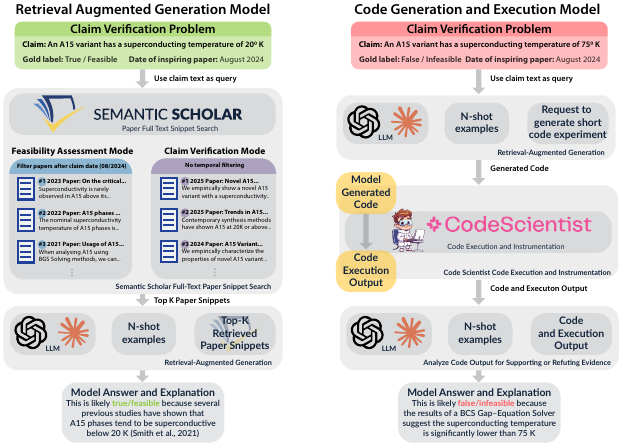}
  \caption{\footnotesize Flow diagrams for two models: the retrieval-augmented generation (\textsc{RAG}) model that retrieves snippets from the full-text of papers using \textsc{SemanticScholar} (left), and a code generation model that executes \textsc{Python} code and examines the output using \textsc{CodeScientist}. }
  \label{fig:model_overview}
\end{figure*}

{\flushleft\textbf{Domain Expert Validation:}} To measure the quality of the claim generation process, a domain expert with a graduate degree in materials science was given each claim and its source paper, and independently asked to determine the validity of the claim.  This was a challenging task, because the claims span broad areas of materials science that would be unusual for any single individual to have expertise within. The domain expert performed this task for 100 claims from the test set, and initially agreed in 93\% of cases (while noting a further 3\% of claims appeared to not meet criteria, such as explicitly referencing the original source paper).  They were then provided with the LLM-generated labels and explanations, and asked to resolve disagreements (either LLM errors, or human error), noting that nearly all errors were a result of the domain expert missing difficult-to-find evidence on their first attempt, and ultimately reaching 99\% agreement after this resolution process. This empirically suggests that the overall data quality is high (96\% after discounting data not meeting generation criteria). Before resolution, interrater agreement using Cohen's Kappa \cite{cohen1960coefficient} was $\kappa = 0.86$, or strong agreement \cite{mchugh2012interrater}, suggesting the claims are highly objective.

%
%
\begin{table*}[t!]
\centering
\footnotesize
{\setlength{\tabcolsep}{2.5pt}
\begin{tabular}{lc*{8}{C{7mm}}c}
\toprule
\textbf{}                                        &   \textbf{Overall}  & \multicolumn{8}{c}{\textbf{Accuracy by Category}}  & \textbf{Cost} \\
\textbf{Model} & \textbf{Accuracy} & \textbf{True} & \textbf{False} & \textbf{Qual.} & \textbf{Qnt.} & \textbf{Exp.} & \textbf{Code} & \textbf{Ther.} & \textbf{Int.} & \textbf{(per 1k)} \\
\midrule
\rowcolor[HTML]{E3E3E3} 
\textsc{Random Baseline} & \cellcolor[HTML]{ffffff} 0.50 & \cellcolor[HTML]{ffffff} 0.50 & \cellcolor[HTML]{ffffff} 0.50 & \cellcolor[HTML]{ffffff} 0.50 & \cellcolor[HTML]{ffffff} 0.50 & \cellcolor[HTML]{ffffff} 0.50 & \cellcolor[HTML]{ffffff} 0.50 & \cellcolor[HTML]{ffffff} 0.50 & \cellcolor[HTML]{ffffff} 0.50 & \cellcolor[HTML]{ffffff} 0 \\ 
\midrule
\rowcolor[HTML]{E3E3E3} 
\multicolumn{11}{l}{\textsc{gpt-4o-mini}} \\
\midrule
~\textsc{Chain-of-Thought (CoT)} & \cellcolor[HTML]{c0e3c0}0.65 &\cellcolor[HTML]{5095c4}0.89 &\cellcolor[HTML]{ffffff}0.40 &\cellcolor[HTML]{cda5e1}0.70 &\cellcolor[HTML]{eee0f5}0.57 &\cellcolor[HTML]{ffe3ca}0.61 &\cellcolor[HTML]{ffe0c5}0.62 &\cellcolor[HTML]{ffd6b2}0.66 &\cellcolor[HTML]{ffcea3}0.69 &\cellcolor[HTML]{fefbfb}\$1 \\ 
\rowcolor[HTML]{F3F3F3}
~\textsc{CoT} + \textsc{20-shot ICL} & \cellcolor[HTML]{bce1bc}0.66 &\cellcolor[HTML]{dceaf3}0.58 &\cellcolor[HTML]{8fbbd9}0.75 &\cellcolor[HTML]{caa1e0}0.71 &\cellcolor[HTML]{e6d2f0}0.60 &\cellcolor[HTML]{ffdbbc}0.64 &\cellcolor[HTML]{ffdbbc}0.64 &\cellcolor[HTML]{ffd9b7}0.65 &\cellcolor[HTML]{ffc490}0.73 &\cellcolor[HTML]{fef9f9}\$2 \\ 
~\textsc{CoT} + \textsc{20-shot ICL} + \textsc{Reflection} & \cellcolor[HTML]{b7dfb7}0.67 &\cellcolor[HTML]{d7e6f1}0.59 &\cellcolor[HTML]{94bedb}0.74 &\cellcolor[HTML]{caa1e0}0.71 &\cellcolor[HTML]{e6d2f0}0.60 &\cellcolor[HTML]{ffdbbc}0.64 &\cellcolor[HTML]{ffdec1}0.63 &\cellcolor[HTML]{ffd6b2}0.66 &\cellcolor[HTML]{ffc490}0.73 &\cellcolor[HTML]{fdf2f2}\$3 \\ 
\rowcolor[HTML]{F3F3F3}
~\textsc{CoT} + \textsc{ICL} + \textsc{RAG (SemanticScholar)} & \cellcolor[HTML]{b3ddb3}0.68 &\cellcolor[HTML]{70a8cf}0.82 &\cellcolor[HTML]{e9f2f8}0.55 &\cellcolor[HTML]{c394db}0.74 &\cellcolor[HTML]{e3ceef}0.61 &\cellcolor[HTML]{ffd9b7}0.65 &\cellcolor[HTML]{ffd6b2}0.66 &\cellcolor[HTML]{ffcea3}0.69 &\cellcolor[HTML]{ffc28c}0.74 &\cellcolor[HTML]{fdf2f2}\$3 \\ 
~\textsc{CoT} + \textsc{ICL} + \textsc{Code (CodeScientist)} & \cellcolor[HTML]{c4e5c4}0.64 &\cellcolor[HTML]{cee1ef}0.61 &\cellcolor[HTML]{b8d4e7}0.66 &\cellcolor[HTML]{cda5e1}0.70 &\cellcolor[HTML]{f0e5f6}0.56 &\cellcolor[HTML]{ffe5cf}0.60 &\cellcolor[HTML]{ffe5cf}0.60 &\cellcolor[HTML]{ffd9b7}0.65 &\cellcolor[HTML]{ffc99a}0.71 &\cellcolor[HTML]{fceeee}\$4 \\ 
\midrule
\rowcolor[HTML]{E3E3E3} 
\multicolumn{11}{l}{\textsc{o4-mini}} \\
\midrule
~\textsc{Chain-of-Thought (CoT)} & \cellcolor[HTML]{def0de}0.58 &\cellcolor[HTML]{c9deed}0.62 &\cellcolor[HTML]{edf4f9}0.54 &\cellcolor[HTML]{d7b8e7}0.66 &\cellcolor[HTML]{ffffff}0.47 &\cellcolor[HTML]{fffaf6}0.52 &\cellcolor[HTML]{fff0e3}0.56 &\cellcolor[HTML]{ffebd9}0.58 &\cellcolor[HTML]{ffd3ad}0.67 &\cellcolor[HTML]{fceeee}\$4 \\ 
\rowcolor[HTML]{F3F3F3}
~\textsc{CoT} + \textsc{20-shot ICL} & \cellcolor[HTML]{a6d7a6}0.71 &\cellcolor[HTML]{a1c6e0}0.71 &\cellcolor[HTML]{a5c9e1}0.70 &\cellcolor[HTML]{bd8ad8}0.76 &\cellcolor[HTML]{dec5ec}0.63 &\cellcolor[HTML]{ffd1a8}0.68 &\cellcolor[HTML]{ffd3ad}0.67 &\cellcolor[HTML]{ffc795}0.72 &\cellcolor[HTML]{ffbc81}0.76 &\cellcolor[HTML]{f3bfbf}\$15 \\ 
~\textsc{CoT} + \textsc{20-shot ICL} + \textsc{Reflection} & \cellcolor[HTML]{a6d7a6}0.71 &\cellcolor[HTML]{9dc3de}0.72 &\cellcolor[HTML]{aacbe2}0.69 &\cellcolor[HTML]{bd8ad8}0.76 &\cellcolor[HTML]{dec5ec}0.63 &\cellcolor[HTML]{ffd1a8}0.68 &\cellcolor[HTML]{ffd1a8}0.68 &\cellcolor[HTML]{ffc99a}0.71 &\cellcolor[HTML]{ffbc81}0.76 &\cellcolor[HTML]{e67d7e}\$30 \\ 
\rowcolor[HTML]{F3F3F3}
~\textsc{CoT} + \textsc{ICL} + \textsc{RAG (SemanticScholar)} & \cellcolor[HTML]{a6d7a6}0.71 &\cellcolor[HTML]{d2e4f0}0.60 &\cellcolor[HTML]{70a8cf}0.82 &\cellcolor[HTML]{c08fd9}0.75 &\cellcolor[HTML]{d9bce9}0.65 &\cellcolor[HTML]{ffd1a8}0.68 &\cellcolor[HTML]{ffd3ad}0.67 &\cellcolor[HTML]{ffc795}0.72 &\cellcolor[HTML]{ffbc81}0.76 &\cellcolor[HTML]{e98a8b}\$27 \\ 
~\textsc{CoT} + \textsc{ICL} + \textsc{Code (CodeScientist)} & \cellcolor[HTML]{b3ddb3}0.68 &\cellcolor[HTML]{b8d4e7}0.66 &\cellcolor[HTML]{a1c6e0}0.71 &\cellcolor[HTML]{c598dd}0.73 &\cellcolor[HTML]{e1c9ed}0.62 &\cellcolor[HTML]{ffd3ad}0.67 &\cellcolor[HTML]{ffd9b7}0.65 &\cellcolor[HTML]{ffcea3}0.69 &\cellcolor[HTML]{ffc490}0.73 &\cellcolor[HTML]{e36c6c}\$34 \\ 
\midrule
\rowcolor[HTML]{E3E3E3} 
\multicolumn{11}{l}{\textsc{claude-sonnet 3.7}} \\
\midrule
~\textsc{Chain-of-Thought (CoT)} & \cellcolor[HTML]{abd9ab}0.70 &\cellcolor[HTML]{7db0d3}0.79 &\cellcolor[HTML]{cee1ef}0.61 &\cellcolor[HTML]{bd8ad8}0.76 &\cellcolor[HTML]{e1c9ed}0.62 &\cellcolor[HTML]{ffd6b2}0.66 &\cellcolor[HTML]{ffd3ad}0.67 &\cellcolor[HTML]{ffc99a}0.71 &\cellcolor[HTML]{ffba7d}0.77 &\cellcolor[HTML]{f8d8d8}\$9 \\ 
\rowcolor[HTML]{F3F3F3}
~\textsc{CoT} + \textsc{20-shot ICL} & \cellcolor[HTML]{a2d5a2}0.72 &\cellcolor[HTML]{599ac7}0.87 &\cellcolor[HTML]{e5eff6}0.56 &\cellcolor[HTML]{b881d5}0.78 &\cellcolor[HTML]{dec5ec}0.63 &\cellcolor[HTML]{ffd1a8}0.68 &\cellcolor[HTML]{ffd6b2}0.66 &\cellcolor[HTML]{ffc490}0.73 &\cellcolor[HTML]{ffb573}0.79 &\cellcolor[HTML]{db4344}\$44 \\ 
~\textsc{CoT} + \textsc{20-shot ICL} + \textsc{Reflection} & \cellcolor[HTML]{bce1bc}0.66 &\cellcolor[HTML]{599ac7}0.87 &\cellcolor[HTML]{ffffff}0.45 &\cellcolor[HTML]{caa1e0}0.71 &\cellcolor[HTML]{ebdcf3}0.58 &\cellcolor[HTML]{ffdbbc}0.64 &\cellcolor[HTML]{ffe5cf}0.60 &\cellcolor[HTML]{ffd3ad}0.67 &\cellcolor[HTML]{ffc795}0.72 &\cellcolor[HTML]{d62728}\$87 \\ 
\rowcolor[HTML]{F3F3F3}
~\textsc{CoT} + \textsc{ICL} + \textsc{RAG (SemanticScholar)} & \cellcolor[HTML]{a6d7a6}0.71 &\cellcolor[HTML]{c1d9ea}0.64 &\cellcolor[HTML]{86b5d6}0.77 &\cellcolor[HTML]{bd8ad8}0.76 &\cellcolor[HTML]{dec5ec}0.63 &\cellcolor[HTML]{ffcea3}0.69 &\cellcolor[HTML]{ffdec1}0.63 &\cellcolor[HTML]{ffc28c}0.74 &\cellcolor[HTML]{ffba7d}0.77 &\cellcolor[HTML]{d62728}\$76 \\ 
~\textsc{CoT} + \textsc{ICL} + \textsc{Code (CodeScientist)} & \cellcolor[HTML]{c8e6c8}0.63 &\cellcolor[HTML]{8fbbd9}0.75 &\cellcolor[HTML]{ffffff}0.50 &\cellcolor[HTML]{d7b8e7}0.66 &\cellcolor[HTML]{ebdcf3}0.58 &\cellcolor[HTML]{ffe3ca}0.61 &\cellcolor[HTML]{ffe5cf}0.60 &\cellcolor[HTML]{ffdec1}0.63 &\cellcolor[HTML]{ffd6b2}0.66 &\cellcolor[HTML]{d62728}\$173 \\ 
\bottomrule
\end{tabular}}

\caption{Model performance on the \textbf{feasibility assessment} task, including overall performance, as well as performance broken down by specific categories of feasibility assessment claim problems.  \textit{True} and \textit{False} represent performance on problems with those gold labels.  \textit{Qual.} and \textit{Quant.} represent performance on qualitative and quantitative problems.  \textit{Exp.}, \textit{Code}, \textit{Ther.}, and \textit{Int.} represent performance on claims focusing on experimental, code/simulation, theoretical, or integrative results, respectively.  Cost represents the estimated model cost per 1000 claims, in US dollars.}
\label{tab:results}
\end{table*}

%
%
\begin{table}[t!]
\centering
\footnotesize
{\setlength{\tabcolsep}{2.5pt}
\begin{tabular}{lc}
\toprule
\textbf{}                                        &   \textbf{Overall}  \\
\textbf{Model}                                   &   \textbf{Accuracy}  \\
\midrule
\rowcolor[HTML]{E3E3E3} 
\multicolumn{2}{l}{\textsc{gpt-4o-mini}} \\
\midrule
~\textsc{RAG (SemanticScholar (No Date))} & \cellcolor[HTML]{91cd91}0.76  \\ 
\rowcolor[HTML]{F3F3F3}
~\textsc{Oracle Source Paper} & \cellcolor[HTML]{a6d7a6}0.71  \\ 
\midrule
\rowcolor[HTML]{E3E3E3} 
\multicolumn{2}{l}{\textsc{o4-mini}} \\
\midrule
~\textsc{RAG (SemanticScholar (No Date))} & \cellcolor[HTML]{56b356}0.90  \\ 
\rowcolor[HTML]{F3F3F3}
~\textsc{Oracle Source Paper} & \cellcolor[HTML]{3da73d}0.96  \\ 
\midrule
\rowcolor[HTML]{E3E3E3} 
\multicolumn{2}{l}{\textsc{claude-sonnet 3.7}} \\
\midrule
~\textsc{RAG (SemanticScholar (No Date))} & \cellcolor[HTML]{63b963}0.87  \\ 
\rowcolor[HTML]{F3F3F3}
~\textsc{Oracle Source Paper} & \cellcolor[HTML]{2ca02c}1.00  \\ 
\midrule
\rowcolor[HTML]{E3E3E3} 
\multicolumn{2}{l}{\textsc{Human Domain Expert}} \\
\midrule
\textsc{Initial Assessment} & \cellcolor[HTML]{49ad49} 0.93 \\
\rowcolor[HTML]{F3F3F3}
\textsc{After Resolving Disagreements} & \cellcolor[HTML]{2ca02c} 0.99 \\
\bottomrule
\end{tabular}}
\caption{Model performance on the \textbf{claim verification} task, using oracle models. $\dagger$Note that due to the high model cost, the \textsc{Oracle Source Paper (Sonnet)} model is assessed on a subset of the test set.
\label{tab:results-claim-verification}}
\end{table}

\section{Baseline Models}

We evaluate performance on the \textsc{Matter-of-Fact} dataset using a selection of baseline models described below.  Models are provided with the text of the claim, and must predict a binary label (true/feasible, or false/infeasible), as well as provide a brief explanation for their reasoning.  All models investigated in this work use in-context learning (\textsc{ICL}), and are characterized across three common base models at different price/performance points, including \textsc{GPT-4o-mini}, \textsc{o4-mini}, and \textsc{Claude Sonnet 3.7}. Our retrieval-augmented generation and code-generation models are shown in Figure~\ref{fig:model_overview}.

\subsection{Feasibility Assessment Models}

{\flushleft{\textbf{Chain-of-Thought (\textsc{CoT}), ICL, Reflection:}}} The language model is provided with a prompt that includes the claim, and a request to think and/or plan before responding in the style of Chain-of-Thought~\cite{10.5555/3600270.3602070}. We also include two variations of this model. The first includes a \textit{20-shot} in-context learning example \cite{brown2020language}, using 20 claim problems (together with their supporting facts and explanations) drawn from the training set, including balanced numbers of true and false claims.  The second includes adding a reflection step \cite{madaan2023self} where, after the initial generation, the model then reflects on its response, then provides a final answer and explanation for the reasoning behind that answer. 

{\flushleft{\textbf{Retrieval Augmented Generation (\textsc{RAG}):}}} Using the claim text as a query, the model first retrieves the \textit{top K} matching full-text snippets from scholarly scientific articles using the \textsc{SemanticScholar} API \cite{kinney2023semantic}, where each snippet generally takes the form of a span of text (approximately 500 words in length) from an article indexed by \textsc{SemanticScholar} that most closely matches the query. To prevent temporal contamination with oracle knowledge, full-text snippets are filtered such that papers authored after the source paper for a given claim are not included in the search.  For example, if a claim was derived from a paper first published on Arxiv in March 2024, then only papers authored in February 2024 or before will be included in the snippet search. The top 20 matching full-text snippets (sorted by the provided relevance score) are included in the language model prompt, in a retrieval-augmented-generation paradigm \cite{lewis2020retrieval}. The prompt for this model also includes a 20-shot \textsc{ICL} example, and request for chain-of-thought reasoning.

{\flushleft{\textbf{Code Generation (\textsc{CodeScientist}):}}} This model is performed in two stages. During the first stage, the model is provided with the claim text, and prompted to generate a code-based experiment or simulation in \textsc{Python} that would produce useful evidence in supporting or refuting the claim.  The code is then executed, and the code and execution results are provided to a second prompt with a request to generate an answer for the feasibility task as well as a supporting explanation.  For code execution, we use the experiment execution portion of \textsc{CodeScientist} \cite{jansen2025codescientist}, which allows executing arbitrary \textsc{Python} code in a virtual sandbox on \textsc{Modal.com}, and supports installing external supporting libraries through \text{PIP}.  While this execution pipeline stores and saves output streams (e.g. \textsc{stdout/stderr}), the model explicitly prompted to save a log of its work, as well as a final list of results, which are then provided back to the model to help make its final decision.  For tractability, we run \textsc{CodeScientist} in a \textit{highly limited form} due to its high overall cost (initially reported as \$4 per experiment), which would be intractable for the size of our dataset ($\approx \$16k$ for \textsc{4k} test claims).  Instead of 25 debug iterations, we run \textsc{CodeScientist} for a single iteration (without reflection), and reduce the experiment time limit from 6 hours to 10 minutes (or 31 total CPU-days across all test claims).  The model is made aware of these limitations in the code generation prompt, and encouraged to design appropriately-scoped experiments and output to support the decision process. 

\subsection{Claim Verification Models}

As a method of characterizing model performance when oracle information is available, Table~\ref{tab:results-claim-verification} also provides two models that perform a \textit{claim verification} task rather than the \textit{feasibility assessment} task -- that is, they do not have the same temporal restrictions, and are able to use data available after the source claim was authored. 

{\flushleft{\textbf{RAG (Temporally Unrestricted):}}} The retrieval-augmented generation model described above, but without temporal restrictions.  For a given claim, snippets from any scientific article may be retrieved, including (potentially) the source article of the claim, or those that cite the source article.

{\flushleft{\textbf{Oracle Baseline:}}} The language model is provided both with the claim, as well as the original source paper the claim was derived from (in the form of the paper's original \LaTeX ~source retrieved from Arxiv) in a retrieval-augmented-generation paradigm.  This baseline measures how well a model can verify the claim when provided with a source scientific article that directly speaks to that claim's validity/feasibility. 

{\flushleft{\textbf{Oracle (Human Domain Expert):}}} The domain expert evaluation, as described in Section~\ref{sec:dataset}.

\subsection{Results}

{\flushleft{\textbf{Feasibility Assessment Results:}}} The performance of all models when evaluated in the \textit{feasibility assessment} mode is shown in Table~\ref{tab:results}.  Performance across all models ranges from \textsc{0.58} to \textsc{0.72}, with the models that use the smallest (and least expensive) base model (\textsc{GPT-4o-mini}) generally performing about 5 percent lower than the two more performant (and more costly) base models, \textsc{o4-mini} and \textsc{Claude Sonnet 3.7}.  Across models, adding features (such as in-context learning, reflection, \textsc{RAG} over \textsc{SemanticScholar}, or \text{Code Generation}) generally provides modest performance improvements, or does not improve performance over the \textsc{Chain-of-Thought} baseline, highlighting the difficulty of this task when using conventional solution methods, and its suitability as a challenge task.  When examining performance broken down by category, we observe that while the overall performance of a given base model is similar with different features, some models are more performant at identifying true/feasible claims than they are at identifying false/infeasible claims, and vice versa.  All models perform better at assessing the feasibility of \textit{qualitative} claims than \textit{quantitative} claims, with this difference between 11\% and 19\% across all models, potentially a result of quantitative claims requiring the ability (through code or other means) of verifying the feasibility of specific numerical values present in the claims.  In line with this reasoning, claims that are based on \textit{experiments} or \textit{code/simulations} consistently achieve the lowest performance, while those based on \textit{theoretical results} are next-highest, with \textit{integrative} claims achieving the highest performance.

%
%
\begin{table*}[t!]
\centering
\footnotesize
\begin{tabular}{lccccc}
\toprule
\textbf{}           &   \textbf{Knowledge}      &   \textbf{Accuracy}         &   \textbf{Accuracy}         & \textbf{Accuracy} &   \textbf{\# Samples} \\
\textbf{Base Model} &   \textbf{Cutoff Date}    &   \textbf{(before cutoff)}  &   \textbf{(after cutoff)}   & \textbf{$\Delta$} &   
\textbf{(before/after)} \\
\midrule
\textsc{gpt-4o-mini}&   Sept 2023               &   0.661                     &   0.664                     & 0.003             &   3236 / 5124    \\
\textsc{o4-mini}    &   May 2024                &   0.694                     &   0.705                     & -0.011            &   5168 / 3192    \\
\textsc{Claude-Sonnet-3-7} &   Oct 2024         &  0.672                      &   0.661                     & 0.011             &   6130 / 2230    \\
\bottomrule
\end{tabular}
\vspace{-1mm}

\caption{\footnotesize Knowledge contamination analysis of base models. In this analysis, performance of the \textsc{Chain-of-Thought + 20-Shot ICL + Reflection} model is shown for claims from papers that were authored before or after a given model's advertised knowledge cutoff date.  Given the temporal nature of the dataset, all \textsc{8.4k} claims across train, development, and test sets were included.  All models show almost identical performance ($\pm1\%$) when tested on claims from papers before or after their knowledge cutoff date, suggesting that knowledge contamination does not play a significant role in performance.
\label{tab:knowledgeanalysis}}
\vspace{-1mm}
\end{table*}

{\flushleft{\textbf{Claim Verification Results:}}} The performance of models when evaluated in the \textit{claim verification} mode is shown in Table~\ref{tab:results-claim-verification}.  In this mode the models have no temporal restrictions, and may use knowledge from the source paper, or papers authored after the source paper (including those that may cite the source paper) as evidence to perform the claim verification task.  These experiments serve two purposes.  First, they identify an effective ceiling of how well a given base model can perform even when provided with the original source article used to create a claim, with \textsc{o4-mini} and \textsc{Claude Sonnet 3.7} capable of achieving nearly a 100\% ceiling performance, while \textsc{gpt-4o-mini} has more modest performance ceiling between 0.71 and 0.76.  Second, these models serve as a consistency evaluation for the claim generation protocol, emphasizing that when strong models are asked to verify the labels of these automatically generated claims, they nearly always agree with the gold label.  Further emphasizing this is the domain expert performance, who (when provided with the original source article), agreed with the LLM-generated label for 99\% of claims after resolving disagreements. 

Taken together, these results empirically demonstrate the generation quality of the feasibility claims, while also emphasizing that common models and architectures still achieve overall modest performance on the feasibility assessment task.  

\section{Discussion}

\subsection{Controlling Potential Confounds}

{\flushleft\textbf{Base-Model Contamination:}} A central part of the framing of our \textit{feasibility assessment} task as a temporally-filtered \textit{claim verification} task is that it requires models to have a minimum of contamination with knowledge beyond a given claim's knowledge cutoff date.  While it is possible that techniques such as model editing and machine unlearning \cite{bourtoule2021machine,tarun2023fast,liu2025rethinking} may eventually allow the knowledge in a base model to be temporally filtered to minimize this contamination, this may have limited success in current forms \cite{lynch2024eight,deeb2024unlearning,du2024textual}.  Instead, here we aim to measure how much of the current model performance is likely due to model contamination (from, for example, the base model being trained on the source articles used to generate the claims).  To measure this, we examine each base model's performance 
for claims extracted from papers before and after the base model's advertised training data knowledge cut-off dates.  The results, shown in Table~\ref{tab:knowledgeanalysis}, show that the performance of the base models on claims from papers authored after their knowledge cutoff is \textit{nearly identical} to the performance on claims authored by papers that are before the knowledge cutoff date.  
This empirically suggests that the performance of current base models on the feasibility assessment task is not due to model contamination, but due to other properties, such as their capacity for reasoning.

{\flushleft\textbf{Collecting Claims from Scientific Papers:}}
In this work, the claims we use for the feasibility assessment task (and the associated hypotheses underlying those claims) are collected from scientific articles.  It is common that the narrative presented in an article differs from the actual scientific process that was undertaken, which typically includes failed experiments, promising initial directions that turned out to be dead ends, and other complexities that are often left out when writing a paper.  As such, the claims (and associated hypotheses) extracted from papers may have a different (and more polished) character than those one naturally explores during the scientific discovery process.  While it is currently difficult to quantify this potential difference, we wish to acknowledge that it may exist, and that this may ultimately affect transfer performance in models trained or evaluated on literature-derived claims (like those in \textsc{Matter-of-Fact}) to real-world discovery scenarios.

\subsection{Performance Characterization}

{\flushleft\textbf{Pragmatic Ceiling Performance:}} While we empirically show that the feasibility of many claims can be assessed using inexpensive means, the models we demonstrate are far from achieving perfect performance on this task.  Pragmatically, a model that achieves near 100\% performance would be able to (with near perfect accuracy) determine whether claims are likely to be feasible or infeasible through some combination of literature search, inexpensive code-based experimentation, world modeling, and other means.  Achieving 100\% performance is likely impractical, as many scientific claims can only be verified with empirical work, and not with literature search or simulation, particularly for those (most impactful) scientific results that are surprising because they run counter to expectations. That being said, even though effective ceiling performance on feasibility assessment tasks is likely to be less than 100\%, increasing model performance on this task even a modest amount can have practical utility for improving the efficiency of discovery systems. As we show in \textsc{Appendix}~\ref{sec:appendix_utility}, for a hypothetical hypothesis generation system where 1\% of the hypotheses are true, the performance of our current-best model could potentially allow discovering 60\% of the true hypotheses while reducing experiment costs by 80\% -- a large overall budget reduction, at the cost of reducing the recall of finding true hypotheses by approximately 40\%. 

{\flushleft\textbf{Limitations in Baseline Models (Retrieval):}}
It is entirely plausible that assessing the feasibility of complex state-of-the-art scientific claims would require integrating knowledge that comes from more than one scientific article.  While our retrieval-augmented-generation system presents the \textsc{top-K} paper snippets to the model making the feasibility assessment, in our baseline system these snippets come from a single search query, and multiple search queries may be required to collect different types of evidence that, together, could be integrated to improve the feasibility assessment.  An initial pilot system that we constructed that iteratively allowed up to 5 rounds of evidence collection (each with a different query) before making an assessment did not appear to improve feasibility performance, suggesting integrating this knowledge into improved feasibility task performance is non-trivial.  Similarly, it is plausible that building a specialized domain-specific corpus of supporting scientific articles may improve the utility of the retrieved knowledge, and increase task performance.

{\flushleft\textbf{Limitations in Baseline Models (Code):}}
Due to the current high estimated cost in running code generation systems, and the large size of the \textsc{Matter-of-Fact} test set, our code generation baseline was run in a highly limited form (i.e. short runtime, no debug iterations) that is best considered an approximate measure of \textit{zero-shot code generation performance} on this feasibility assessment task, without the benefits of debugging iterations or the long experiment runtimes that code-based solutions to this task would almost certainly require.  Some of these limitations are pragmatic, like cost and runtime, and are rapidly reduced as (for example) open language models for code generation that can be run on local hardware begin to approach paid \textsc{API-based} model performance.  However, we ultimately believe that progress in code-based experimentation will require building systems that integrate materials science domain tools, software, and databases. We did not include any materials-science specific tooling in \textsc{CodeScientist}'s code retrieval library and required it to instead rely on the base model weights to implement these complex tool interfaces.  It is highly likely that near-term improvement on this task will require a degree of manually integrating this tooling, just as other scientific agents (such as \textsc{Biome}~\cite{huang2025biomni} in the biomedical domain) are currently using hand-build interfaces to domain-specific tools and databases to support their discovery tasks.

\section{Conclusion}
We present \textsc{Matter-of-Fact}, a benchmark for assessing the feasibility of \textsc{8.4k} scientific claims in four high-impact subdomains of materials science: superconductors, semi-conductors, batteries, and aerospace materials.  We frame the feasibility assessment task as a temporally-filtered claim verification task, and empirically demonstrate that  strong baseline models using a variety of solving methods (including literature search and code generation) reach only modest performance on this task (72\%).  Performance on feasibility assessment can directly translate to improving automated scientific discovery systems, particularly in hypothesis generation, where filtering infeasible hypotheses can make scientific discovery more efficient, and lower overall experiment costs.  Ultimate solution methods for the feasibility assessment task are likely to require a combination of reasoning over deep literature search, code-based simulation, and world modeling at the scale of subdomains.

\section*{Limitations}

{\flushleft\textbf{Temporal Filtering for Prediction:}} Temporal datasets that use the notion of backtesting offer the opportunity to construct prediction datasets for high-impact domains \citep[e.g.][link prediction for cancer biology]{luo-etal-2018-scientific} where the knowledge a system is predicting is potentially beyond current human knowledge, and for which gold labels are infeasible to construct.  Temporal filtering assumes well-controlled models that have not been contaminated with data past their temporal filtering date.  In this work we characterize the contamination rate of our base language models, and this analysis suggests that data contamination either does not exist, or is not a significant factor driving current performance. That being said, users of this benchmark should characterize the performance of novel base models to characterize how much data contamination may play a role. 

{\flushleft\textbf{Cost-Benefit Analyses:}} Pragmatically, to be useful for filtering scientific hypotheses, feasibility assessment methods must be able to perform well at scale.  This necessitates that any pilot experiments (including code-based simulations) must be fast and inexpensive to run, otherwise the feasibility assessment step may be impractically expensive to provide overall cost savings.  That being said, different applications and end-users may have varying preferred cost/performance points, and we encourage  reporting performance as a function of overall cost (as we have done in this work) to help accurately assess the cost vs benefit of proposed models.  It is our hope that providing a large-scale benchmark that necessitates developing inexpensive feasibility assessment methods will help facilitate innovation in this direction.

\section*{Acknowledgments}
This research was developed with funding from the Defense Advanced Research Projects Agency's (DARPA) SciFy program (Agreement No. HR00112520300) to PJ at the University of Arizona. The views expressed are those of the author and do not reflect the official policy or position of the Department of Defense or the U.S. Government. PJ has an outside interest in the Allen Institute for Artificial Intelligence.  This interest has been disclosed to the University of Arizona and reviewed in accordance with its conflict of interest policies. We thank the members of the \textsc{DARPA} Scientific Feasibility (SciFy) program for thoughtful discussions. We also wish to thank the anonymous reviewers for their helpful comments, and their particular observation that literature-derived claims may be more polished than those present in earlier stages of the discovery process.

\bibliography{custom,anthology}

\appendix
\section{Utility for Hypothesis Filtering}
\label{sec:appendix_utility}

%
%
\begin{figure}[h]
  \centering
  \includegraphics[scale=0.50]{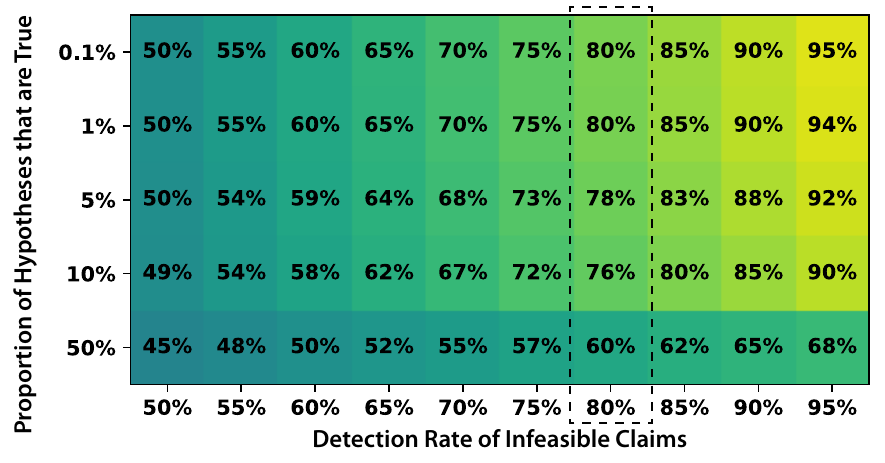}
  \caption{\footnotesize Relative efficiency (in terms of reduction in the number of experiments needed to be run) for a hypothetical automated scientific discovery (ASD) system that generates hypotheses with a certain true positive rate \textit{(y-axis)}, after those hypotheses have been pre-filtered by a feasibility assessment system such as the models described in this work.  This plot assumes a true positive (i.e. \textit{feasible}) detection rate of 0.60, corresponding to the \textsc{RAG (SemanticScholar, o4-mini)} model in Table~\ref{tab:results}, while the highlighted region corresponds to that model's infeasible claim detection rate (82\%).  For a hypothetical ASD system where 1\% of the hypotheses it generated were \textit{true/feasible}, the \textsc{RAG} model would reduce the number of experiments (i.e. cost) by 80\%, while still discovering 60\% of the true hypotheses.}
  \label{fig:detection_efficiency}
\end{figure}

Feasibility assessment has utility in impactful tasks such as (semi-automated) scientific discovery, particularly in the context of hypothesis generation.  Hypothesis generation systems \citep[e.g][]{lu2024ai,jansen2025codescientist,o2025sparks} have the capacity to generate an impractically large number of possible hypotheses (framed as claims) that one could test, and as a result running all their proposed experiments is costly (at best) and intractable (at worst).  Coupling hypothesis generation with feasibility assessment would allow filtering out hypotheses that are unlikely to be feasible -- i.e. yield experimental results that support the hypothesis -- and ultimately increase the efficiency of scientific discovery systems in terms of the number of positive discoveries that can be made on a given budget.  In automated hypothesis generation where overall likelihood of a hypothesis yielding positive results is low, increasing efficiency is dominated by correctly identifying (and filtering) infeasible hypotheses/claims.  Figure~\ref{fig:detection_efficiency} shows a plot of experiment efficiency (in terms of the reduction in the number of experiments that would need to be run) for hypothetical hypothesis generation systems that have different rates of generating true hypotheses, with the performance of the best-performing model (\textsc{RAG (Semantic Scholar)} using \textsc{o4-mini}) highlighted.  For a hypothetical hypothesis generation system where 1\% of its hypotheses are true, this model would reduce the number of experiments needed to be run by approximately 80\%, while still discovering 60\% of the true hypotheses.  This highlights that even systems with middle performance can have practical utility (in terms of cost savings) when coupled with scientific discovery systems. 

\end{document}